# Wavelet based inpainting detection*

## Image inpainting detection based on Dual-Tree Complex wavelet, semantic segmentation and noise inconsistencies

Barglazan Adrian-Alin[1], Brad Remus Ovidiu[1]

Department of Computers and Electrical Engineering, Faculty of Engineering, University "Lucian Blaga", Sibiu, Romania[1]

adrian.barglazan@ulbsibiu.ro, remus.brad@ulbsibiu.ro

***With the advancement in image editing tools, manipulating digital images has become alarmingly easy. Inpainting, which is used to remove objects or fill in parts of an image, serves as a powerful tool for both image restoration and forgery. This paper introduces a novel approach for detecting image inpainting forgeries by combining DT-CWT with Hierarchical Feature segmentation and with noise inconsistency analysis. The DT-CWT offers several advantages for this task, including inherent shift-invariance, which makes it robust to minor manipulations during the inpainting process, and directional selectivity, which helps capture subtle artifacts introduced by inpainting in specific frequency bands and orientations. By first applying color image segmentation and then analyzing for each segment, noise inconsistency obtained via DT-CW we can identify patterns indicative of inpainting forgeries. The proposed method is evaluated on a benchmark dataset created for this purpose and is compared with existing forgery detection techniques. Our approach demonstrates superior results compared with SOTA in detecting inpainted images.***

### I. Introduction

Image inpainting is the process of reconstructing lost or deteriorated parts of an image. It has evolved significantly over the past decades. It is an essential task in various applications such as image restoration, content removal, and computer graphics. Traditional inpainting techniques have given way to more sophisticated methods powered by advances in deep learning, particularly Convolutional Neural Networks (CNNs), Transformer models, and diffusion-based approaches. This review provides an overview of these contemporary methods, highlighting their strengths and challenges. Before the rise of deep learning, inpainting relied on traditional approaches like patch-based methods and diffusion-based methods. Patch-based techniques like the ones suggested by [1] operate by copying and pasting similar patches from the known regions of the image to fill in the missing areas. Diffusion methods [2] progressively propagate information from the known regions towards the unknown, gradually refining the inpainted content. While these techniques achieved some success, they often struggled with complex missing regions, leading to blurry or repetitive results. More newer methods like [3] introduces a novel approach to image inpainting that leverages contextual attention mechanisms to fill in large missing regions in images. Traditional convolutional networks struggle with inpainting due to their limited ability to utilize information from distant spatial locations. The proposed method addresses this by explicitly borrowing information from surrounding image areas through a contextual attention layer. This layer helps synthesize new image structures and textures that are consistent with the existing parts of the image. The approach is validated on multiple datasets, including CelebA, DTD, and ImageNet, demonstrating superior results compared to existing methods. Even more recent research from 2021 [4] explores image inpainting by employing diffusion models in a latent space. By focusing on latent representations instead of pixel space, the model efficiently learns and restores missing parts of images while preserving the global structure and ensuring high-quality outputs. This method leverages the strengths of both latent space manipulation and diffusion processes to achieve significant improvements in the visual fidelity of inpainted images. In the same category – as diffusion methods, the [5] paper presents a way for high-resolution image synthesis using latent diffusion models. Diffusion models have been successful in generating high-quality images by iteratively denoising samples from a Gaussian distribution. The latent diffusion approach improves efficiency and scalability by operating in a latent space rather than directly on pixel space. This allows for generating high-resolution images with lower computational costs. The method achieves state-of-the-art results in various image synthesis tasks, highlighting its potential for practical applications. While the advancements in image inpainting and object removal technologies offer numerous benefits, they also bring significant ethical challenges. The potential for misuse in the justice system underscores the need for stringent regulations, ethical guidelines, and the development of forensic tools to detect image tampering. Ensuring that these technologies are used responsibly is crucial to prevent their misuse and protect the integrity of visual evidence in legal contexts

Detecting image inpainting forgeries is a crucial task in digital forensics and image authentication. Inpainting forgeries, created by sophisticated methods, can seamlessly remove objects or alter images, making detection challenging. One primary approach to detecting these forgeries involves analyzing inconsistencies at various levels, such as noise, texture, and blur. Noise level analysis is essential because authentic images generally exhibit consistent noise patterns throughout. Inpainted regions, however, often display noise discrepancies, as inpainting algorithms may not replicate the original noise characteristics perfectly. Techniques like the Dual-Tree Complex Wavelet Transform (DT-CWT) [6] are particularly effective in identifying these noise level inconsistencies, capturing subtle noise patterns and highlighting discrepancies between inpainted and authentic regions. Texture analysis is another crucial aspect, as inpainting algorithms might fail to maintain texture continuity, especially in complex scenes. Examining the texture features can reveal abrupt changes or unnatural patterns that suggest tampering. Techniques such as co-occurrence matrices and local binary patterns are commonly used to quantify texture and detect anomalies indicative of inpainting. Blur detection is also significant in identifying inpainted regions. Inpainted areas may exhibit different levels of blur compared to the original parts of the image, as the inpainting process might not perfectly match the sharpness or smoothness of the surrounding areas. Gradient-based methods assess the sharpness and gradient distribution across the image, identifying areas with inconsistent blur levels. Combining these analysis techniques provides a robust framework for detecting inpainting forgeries. Leveraging multi-level feature analysis allows for the identification of subtle inconsistencies that single-level approaches might miss. This holistic approach is crucial for maintaining the integrity and authenticity of digital images in an era where advanced inpainting techniques are increasingly accessible.

## II. Related work

In the world of investigations, ensuring evidence is genuine is crucial. Forensic detection methods come to the rescue, helping identify if documents, images, or videos have been tampered with. These methods aim to confirm the evidence's authenticity and integrity. There are two main approaches active and passive. Active (Watermark-Based), like a hidden security tag, these methods rely on pre-embedded data, such as digital watermarks, within the original content. Verifying the presence and condition of this embedded information helps expose any tampering attempts. Passive (Blind) methods don’t require prior knowledge of the original or embedded information. Instead, it analyzes the content itself for inconsistencies that might indicate manipulation. This is particularly useful for evidence from social media or open-source platforms, where the original might not be available. Usually how passive methods work is by inconsistency analysis combined with statistical analysis. These methods are like detectives examining the scene for clues. They look for inconsistencies within the content itself, like pixel-level differences in images, lighting/shadow inconsistencies, or unnatural pattern repetitions. Statistical techniques analyze the content's properties, like noise patterns or compression artifacts.

Forgeries can introduce statistical anomalies that this analysis can detect. For e.g. different cameras and image acquisition processes introduce unique noise patterns. Forgeries involving splicing or copy-pasting from different sources might exhibit inconsistencies in noise characteristics. Statistical measures like standard deviation or kurtosis can analyze noise patterns in different image regions. Deviations from expected values within the image can indicate potential manipulation. The paper "Image Noise and Digital Image Forensics" by Thibaut Julliand, Vincent Nozick, and Hugues Talbot [7] provides an in-depth exploration of the role of noise in digital image forensics. Noise, an intrinsic element in all forms of imaging, stems from various sources during the image acquisition process, such as the nature of light and optical artifacts, as well as the conversion from electrical signals to digital data. The study highlights several key aspects like sources, models and techniques. Noise Sources and Models can categorize the different types of noise that affect digital images and discusses various models used to represent these noises. Understanding these models is crucial for forensic analysis because noise patterns can be distinctive and unique to specific imaging devices. The paper details how noise can be utilized in digital image forensics to detect forgeries. Techniques like analyzing sensor pattern noise (SPN) and photo-response non-uniformity (PRNU) are explained. These techniques help in identifying the source of an image and detecting any tampering by examining inconsistencies in the noise patterns. The authors in [8] propose using intrinsic sensor noise differences, which often vary due to different ISO settings, to identify tampered regions in images. The technique involves estimating noise levels locally using a PCA-based algorithm and then clustering the weighted noise levels with k-means. Experimental results demonstrate the method's effectiveness in localizing and detecting spliced regions, outperforming several state-of-the-art techniques. The paper [8] introduces a method to detect image splicing by leveraging inconsistencies in sensor noise levels. By analyzing noise patterns unique to camera sensors, the method identifies discrepancies indicating spliced regions. This approach proves effective in revealing forgeries that are not discernible through visual inspection alone. They first estimated locally the noise based on PCA, then estimated noise levels are clustered via k-means. This paper [9] presents an image forgery localization method utilizing a fully convolutional network (FCN) enhanced with noise features. The approach uses noise patterns to highlight subtle changes in images, thereby improving the network's ability to generalize and detect tampered regions. The FCN architecture generates pixel-wise predictions, and the inclusion of a region proposal network enhances robustness. Experimental results on standard datasets demonstrate the method's effectiveness in accurately locating tampered areas, showcasing improved generalization and robustness over existing techniques. A more recent attempt to detect forgeries based on noise was done by authors in [10]. The paper presents a dual-branch image manipulation detection technique that leverages noise and edge features to improve the accuracy of identifying manipulated images. The approach involves two primary branches: Noise Feature Branch **-** This branch focuses on analyzing the noise characteristics of the image. Noise patterns are intrinsic to images captured by digital sensors and are often altered when images are manipulated. By examining these noise inconsistencies, the method can highlight areas that are likely tampered with**.** Edge Feature Branch - this branch analyzes the edges within the image. Manipulation often introduces unnatural edges or alters existing ones. By closely examining edge patterns and inconsistencies, the method can detect signs of image tampering. The paper describes a convolutional neural network (CNN) architecture with two parallel branches: one for noise detection and one for edge detection. Each branch processes the image independently and extracts relevant features. A fusion layer combines the outputs from both branches to make a final decision about the presence of manipulation. This combined approach leverages the strengths of both noise and edge analysis, providing a more robust detection mechanism. In [11] the authors present a novel approach for detecting post-processed image forgeries by utilizing a Signal Noise Separation-based Network (SNIS). This method draws an analogy between image forgery detection and blind signal separation, treating the problem of forgery detection as one of separating signal from noise. The main core idea is to separate the image content (signal) from the noise that is inherently introduced by the image sensor. Since post-processing operations often alter these noise patterns, the method can effectively identify forgeries. The proposed network consists of two main branches: one for noise extraction and one for signal extraction. This dual-branch structure allows the network to learn and distinguish between authentic image content and manipulated regions. The authors trained their model using large datasets of both genuine and forged images, ensuring robust performance across various types of post-processed forgeries. Another automated tool was introduced in [12] designed to detect image forgeries through noise analysis. The approach leverages the inherent noise patterns that digital cameras imprint on images. By analyzing these noise characteristics, the system can identify discrepancies indicative of tampering. The system extracts noise from an image using advanced signal processing techniques. This involves separating the noise components from the actual image content. A background stochastic model is defined to characterize the typical noise patterns produced by a camera sensor. This model serves as a baseline for comparison. The extracted noise is analyzed locally across different regions of the image. The method identifies areas where the noise pattern deviates significantly from the expected model, suggesting potential tampering. Various statistical techniques are employed to quantify the degree of anomaly in noise patterns. These techniques help in distinguishing genuine parts of the image from manipulated ones. The Noisesniffer operates in a fully automatic mode, requiring minimal user intervention. The tool processes the image, extracts noise, analyzes patterns, and provides a decision on the authenticity of the image.

While significant advancements have been made in image forgery detection using noise analysis and machine learning, these methods still face challenges related to computational efficiency, generalization, and robustness against various types of forgeries. To mention a few of the current limitations:

- many methods struggle with low-resolution or highly compressed images
- noise analysis and pattern detection are computationally intensive, requiring significant processing power.
- machine learning-based methods require extensive training datasets to achieve high accuracy
- high sensitivity can lead to false positives, while less sensitivity might result in undetected forgeries
- some methods might produce false positives in images with naturally high noise variations or by different camera models and settings.

## III. Theoretical Aspects

### A. Dual Tree complex wavelet

The Dual Tree Complex Wavelet Transform (DT-CWT) is an image processing method that offers several advantages over traditional Discrete Wavelet Transforms (DWT). DWT decomposes an image into subbands containing information about different spatial frequencies and orientations. However, it treats horizontal and vertical details separately. DT-CWT addresses this by performing the decomposition on two separate trees simultaneously (called A and B tree in the original paper [6] (see below figure). These filter banks share identical frequency responses but have slightly different phases, achieved through specific filter design techniques like Kingsbury Q-shift filters. This redundancy helps in achieving near-shift invariance. Each filter bank employs scaling functions (φ(x)) and wavelet functions (ψ(x)) to decompose the signal at different scales and orientations. Inpainting detection identifies areas within an image that have been altered or filled in.

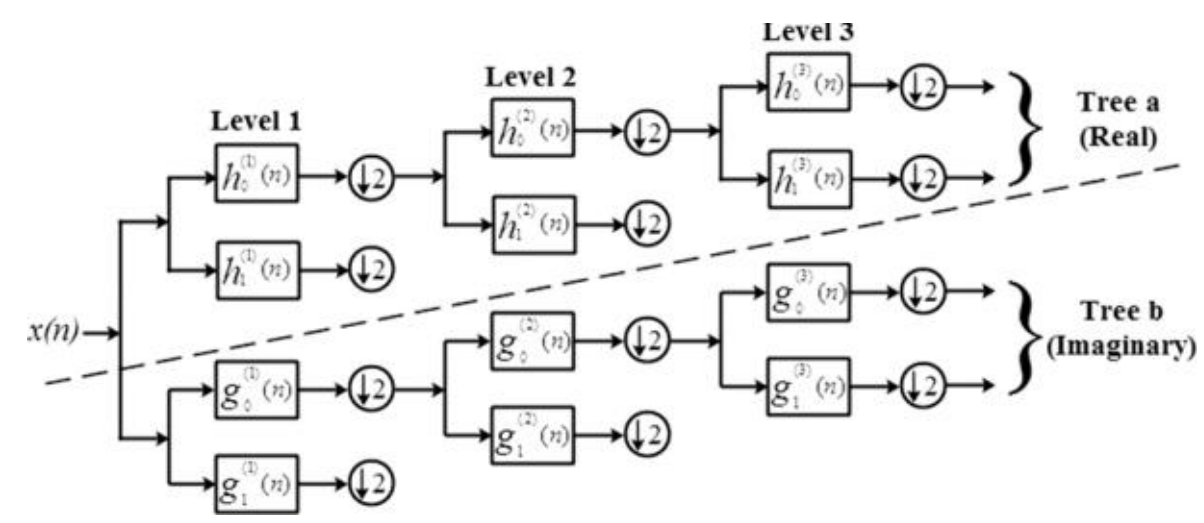

Fig. 1. Representation of dual-tree complex wavelet transform over 3 levels from the original Kingsbury paper

The Dual-Tree Complex Wavelet Transform (DTCWT) is effective as a feature extractor in this context due to several properties. Improved Directionality and Texture Representation: DTCWT captures edge and texture information more effectively than traditional wavelet transforms, making it adept at detecting discrepancies in texture and edge continuity caused by inpainting. Unlike traditional wavelet transforms, DTCWT maintains consistent feature extraction despite shifts, ensuring reliable detection of inpainted regions regardless of their position in the image. The dual-tree structure of DTCWT captures both magnitude and phase information, revealing subtle inconsistencies in inpainted areas that may not be apparent through intensity analysis alone. Phase congruency, derived from phase information, can indicate disruptions in the natural flow of the image's texture or edges. Reduced Aliasing: DTCWT exhibits less aliasing in its subbands compared to traditional wavelet transforms, ensuring cleaner and more representative features, which enhances the detection of inpainted regions, especially those that are small or highly precise. DTCWT's subbands provide a clearer distinction between high-frequency noise components and inpainting artifacts, effectively isolating localized changes in frequency and directional ranges to pinpoint manipulated areas. See the below illustration (all inpainting methods are done using [13] – and source image is from [14] – more details will be provided in the results part):

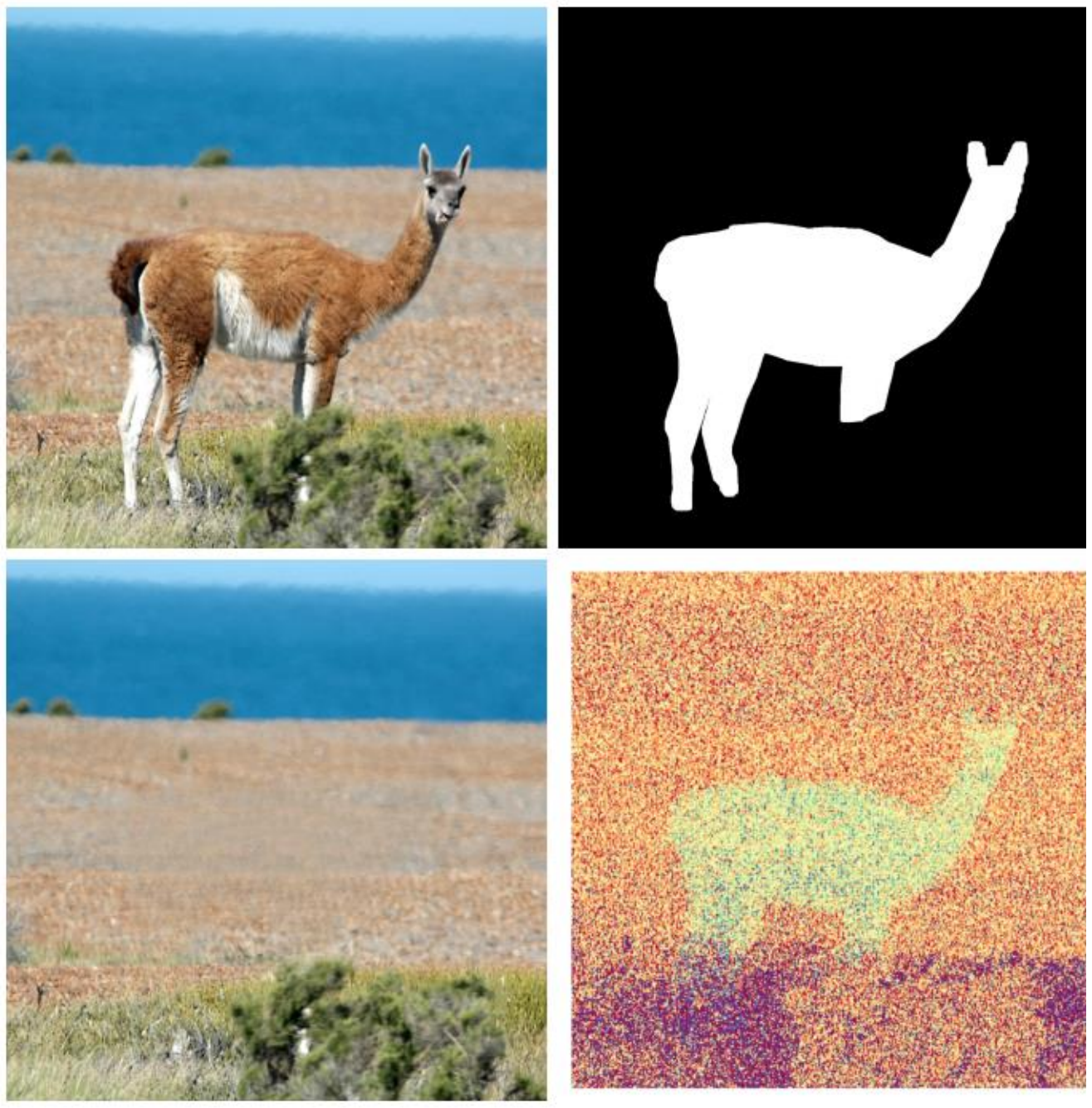

Fig. 2. First row: original image and mask image of the object to be removed. Second row: image inpainted and DTCWT (one level of decomposition selected only one band – third one – real values image enhanced – used spectral coloring and values normalized

In practice only the first level of decomposition shall be applied, all 6 orientations bands are considered (both original and imaginary part) resulting in a total of 12 images that are analyzed.

### *B. Noise level analysis and estimation*

The starting point of the proposed method is the idea expressed in [15]. In the paper, the authors present a novel approach for estimating the noise level in images. The method leverages principal component analysis (PCA) on image patches to identify noise-dominant components. They assume that noise follows a Gaussian distribution. Based on this concept, they applied PCA to separate noise from signal by analyzing eigenvalues. The noisy image I can be modeled as $I = I_0 + N$, where $I_0$ is the clean image and N is the noise. PCA is applied to small patches of the image to decompose the data into principal components. For a set of image patches X, PCA finds the eigenvalues $\lambda_1, \lambda_2, ..., \lambda_n$ of the covariance matrix of X. The smallest eigenvalue $\lambda_{min}$ from the PCA of the image patches is used to estimate the noise variance $\sigma^2$. The rationale is that the smallest eigenvalue is dominated by the noise component due to its isotropic nature. To better analyze the noise inconsistencies, we proposed to alter the above method, and apply first a dual-tree complex wavelet decomposition, and then perform the noise inconsistency inside the HH bands of the dual tree complex wavelet. The above method works well on both natural and synthetic images, but in the case of highly texturized images it struggles to yield a good result. Based on the above work, we postulate that inside an image, if it were to apply color segmentation, and apply the method per segments of same color and texture, then we can extract the noise variance at segment level. Thus, defining a noise estimation at the segment level, we then check if all the patches inside color segmentation are in the same threshold as the noise estimated for the entire segment. In case there is a variance more than 30%-50%, we classify that region as suspicious, and it will enter further analysis. To demonstrate this, several images were selected. Using the above noise estimation level 3 set of noise were computed: from entire image, from the region outside the object removed and from the region where the object was removed. For the simplicity of tests, we've selected images which contains only one type of textures. On the below images, we can see the original image on the left side, the mask image on the right and the altered image on the bottom. Below that there is another image representing the mask applied to the inpainted region – only the inside mask and only the outside mask. In the first instance the noise of the hole altered image is combined. Then the noise is computed for only the inside and outside mask separately. The method [15] is applied to for noise estimation, but instead of applying it directly to RGB image, it is applied to each band (real and imaginary) of the DT-CWTs.

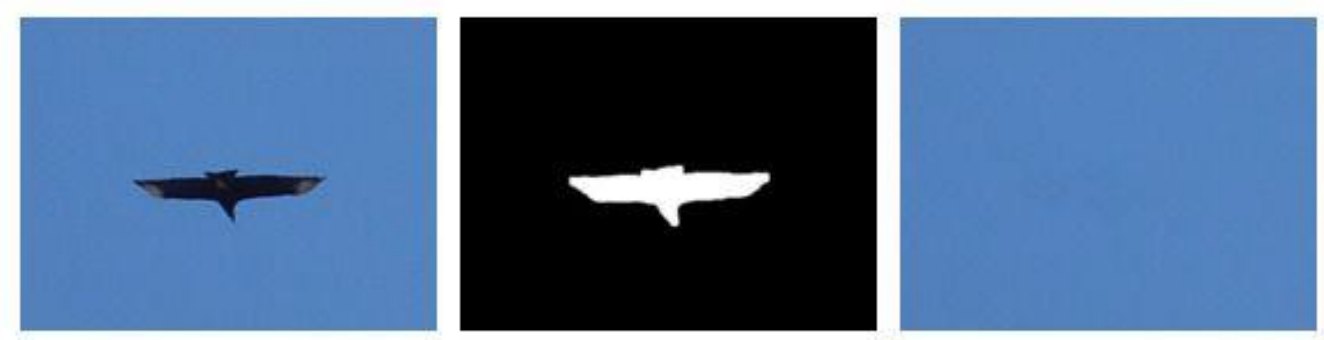

Fig. 3. Original image / Mask image of object to be removed / Altered (inpainted) image

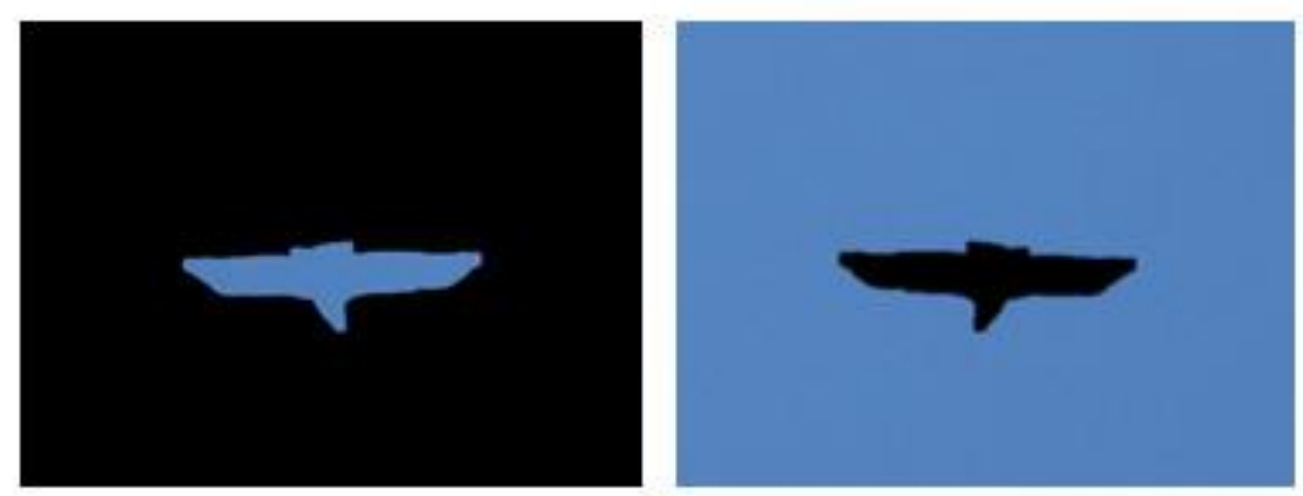

Fig. 4. Image altered image inside mask /outside mask

For the analysis from the complex wavelet decomposition only the real and imaginary were taken. Also, all the bands (0 to 5) are analyzed. The results are highlighted in the table below.

TABLE I. APPLYING NOISE ESTIMATIONS ON DT-CWT BANDS

| ***Band*** | ***Band information (real / imaginary)*** | ***Measurement from:*** | | |
|---|---|---|---|---|
| | | **Mask** | **Outside Mask** | **Entire band (image0** |
| 0 | Real | 205.92 | 278.76 | 276.88 |
| 0 | Abs | 156.84 | 159.88 | 158.88 |
| 1 | Real | 104.37 | 175.08 | 173.60 |
| 1 | Abs | 119.50 | 174.29 | 200.76 |
| 2 | Real | 177.33 | 267.01 | 264.88 |

| ***Band*** | ***Band information (real / imaginary)*** | ***Measurement from:*** | | |
|---|---|---|---|---|
| 2 | Abs | 86.67 | 191.51 | 193.46 |
| 3 | Real | 104.75 | 217.22 | 214.40 |
| 3 | Abs | 115.04 | 151.89 | 148.42 |
| 4 | Real | 120.11 | 237.37 | 231.97 |
| 4 | Abs | 113.70 | 171.00 | 169.79 |
| 5 | Real | 126.02 | 205.27 | 209.28 |
| 5 | Abs | 212.75 | 269.44 | 280.47 |

The observed data indicates that the noise distribution is significantly impacted by the inpainted area. For example, in the image provided, an analysis of the real values across all six levels of band decomposition reveals a noteworthy pattern. Specifically, the noise distribution from the regions outside the masks closely resembles that of the entire image. However, there is approximately a 50% discrepancy between the noise distribution in the altered region compared to the rest of the image. This highlights the distinct alteration in noise characteristics due to inpainting.

## IV. PROPOSED METHOD

The proposed method aims to identify and enhance regions of interest within texture segments and to differentiate between regions with uniform noise distribution and those with significant variations.

In the context of image inpainting, specifically for object removal, the removed object is invariably encompassed within a larger area. Even when the object is situated at the boundary between multiple regions, each section of the removed object will be seamlessly filled with textures from the adjacent areas. This process inherently alters the noise statistical model within each modified color segment. Given these observations, we propose that the forgery detection method should be applied independently to each texture segment. This ensures that the detection is precise and accounts for the unique characteristics of each color segment. Our novel proposed method can be summarized as follows: first, the image undergoes semantic segmentation, resulting in N texture segments, labeled as $S_i$ for i in the range [0..N). Simultaneously, a one-level DT-CWT decomposition is performed on the grayscale image. For each of the 6 bands (where k is between 0 and 5), we extract the real and imaginary parts separately, resulting in 6 $W_{kReal}$ and $W_{kImaginary}$ matrices. Each segment's color serves as a mask for extracting information from the 12 images. We analyze wavelet coefficients, extract patches, and check for inconsistencies to identify potential forgeries. Minimal variations lead to skipping the band information, while significant variations prompt further analysis.

The extracted band data undergoes normalization, which involves a logarithmic transformation to handle negative values, followed by a z-transform to standardize the data. Several enhancement techniques are applied, including mean modified Wiener filtering, median filtering combined with bilateral filtering, Laplacian un-sharpening, and small median filter residue. These techniques were chosen to highlight the variations within the segment. For each enhanced feature, K-means or c-means clustering with two segments/groups is applied. If two distinct segments are identified within the segmented area, the segment with the largest difference in noise is added to the forged image segments. Thus, for an image, we obtain N segments * 12 matrices of wavelet coefficients * 4 different enhancement methods, resulting in 48 N distinct images. The mean value of these images is computed to obtain the inpainted detection mask. Thus, each pixel will represent an intensity. The higher the intensity, the more likely it is that the pixel was detected in several bands (enhancements).

### A. *Region segmentation and dual -complex wavelet band extraction*

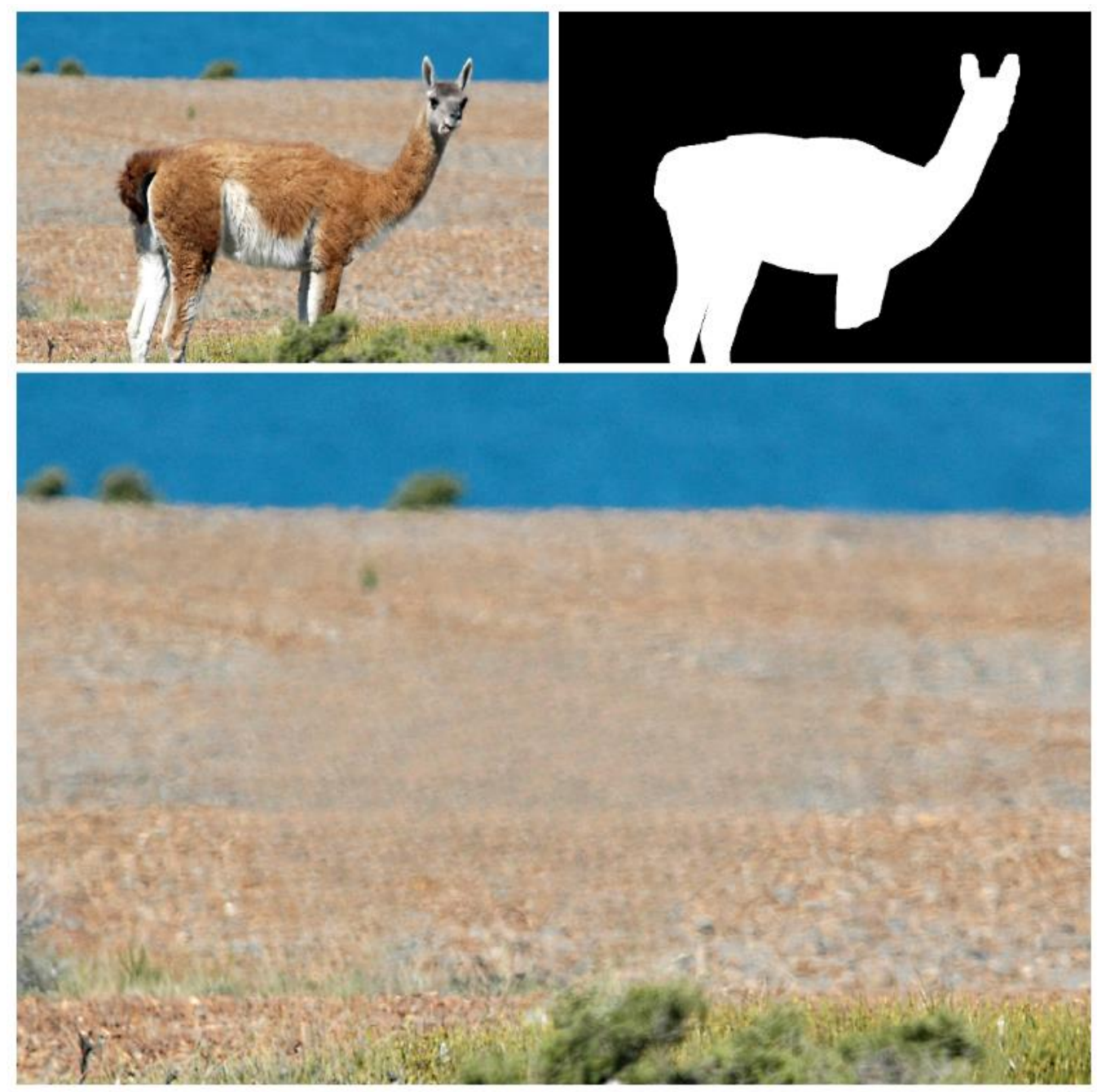

Fig. 5. First row original image / mask image. Second row inpainted image

The above figure on the first row is presenting the original image, followed by mask image of the object to be removed. On the second row, the inpainted image is presented. As mentioned above, the first step is to apply semantic segmentations. For this Hierarchical Feature Selection [16] is employed. Hierarchical feature selection is a technique used to identify important features for machine learning tasks, but with a twist compared to traditional methods. It goes beyond just picking the "best" features and instead builds a hierarchy of feature subsets. Traditional feature selection often treats features independently. Hierarchical selection, however, considers the relationships between features. It creates a hierarchy, like a family tree, where features are grouped based on their similarity and importance. The result of applying HFS on the image is presented in the image below.

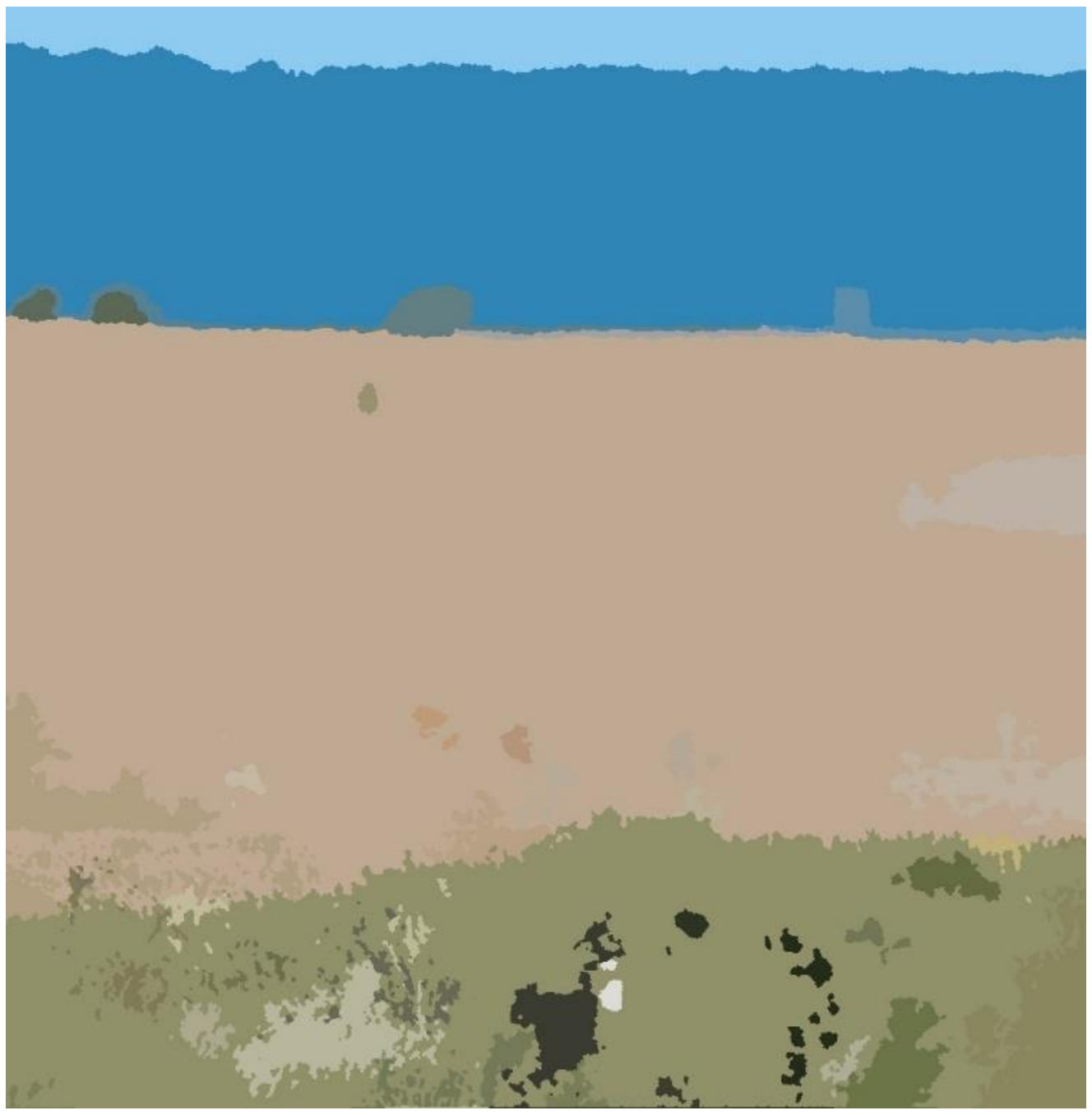

Fig. 6. Output of HFS segmentaton. Each segment has a different color

For the next operation, each individual segment must be considered. For the easiness of the exampled hereby only the largest segment will be further analyzed.

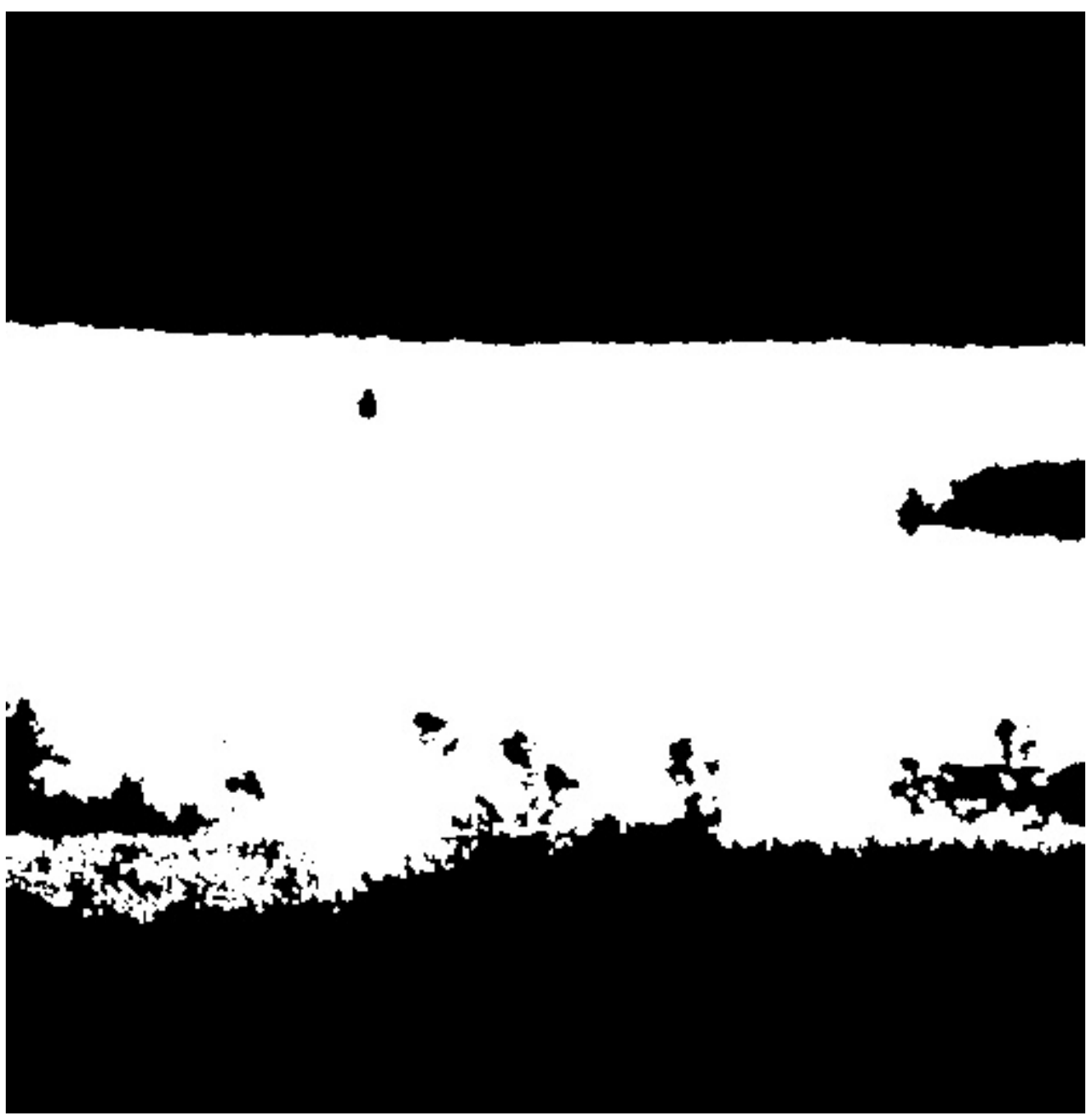

Fig. 7. Segmentation masked further analyzed

The dual-complex tree wavelet is also applied on the entire image. From the analysis only the first decomposition level is taken. From the complex numbers the real and imaginary parts are extracted independently, thus obtaining 12 matrices (images). In the pictures below from first row, from left to right all wavelet coefficients obtained.

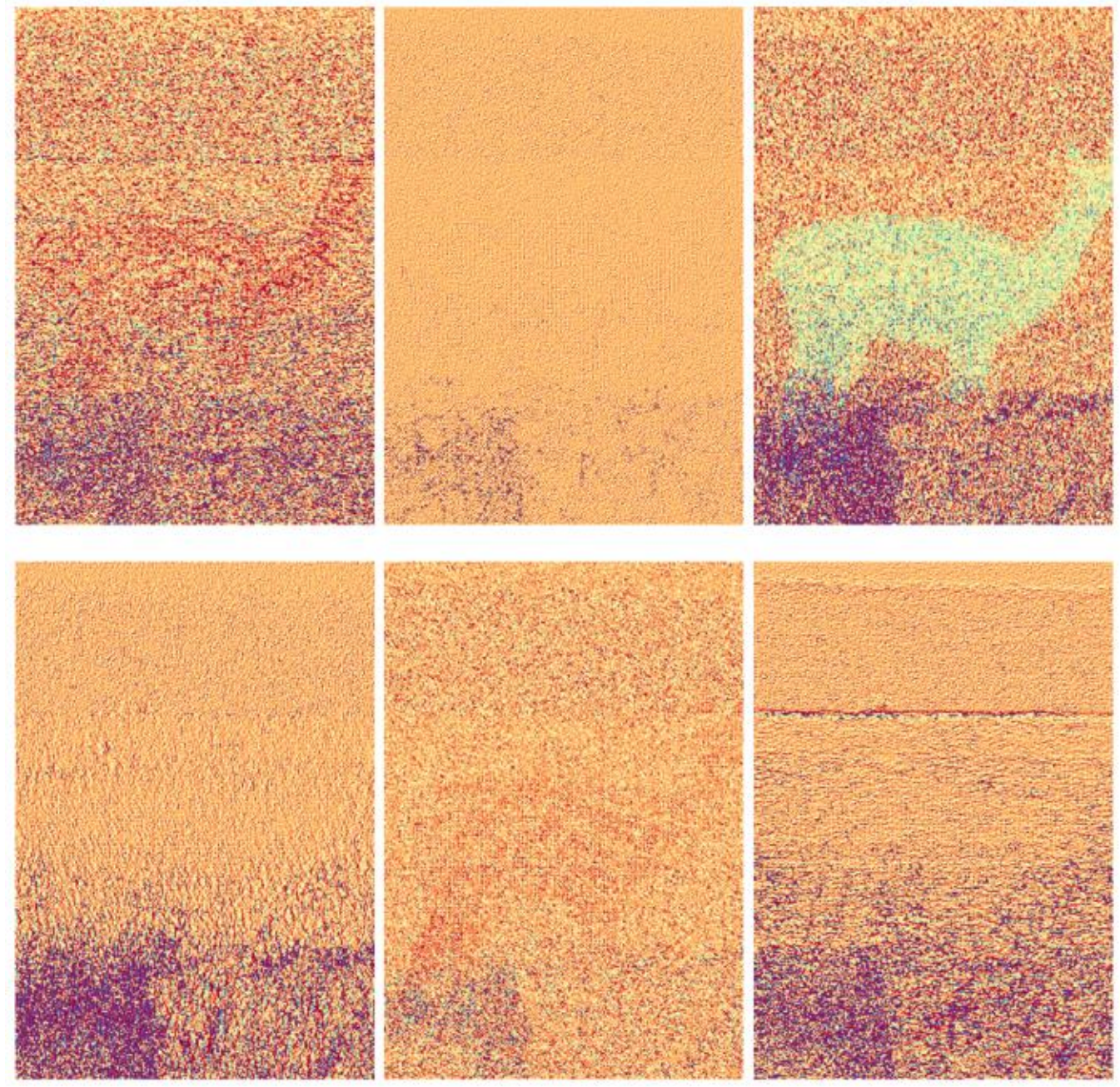

Fig. 8. Real coefficients of DC-WT of the original image. On the first row are the first 3 bands and on the second row the rest of the last 3 bands

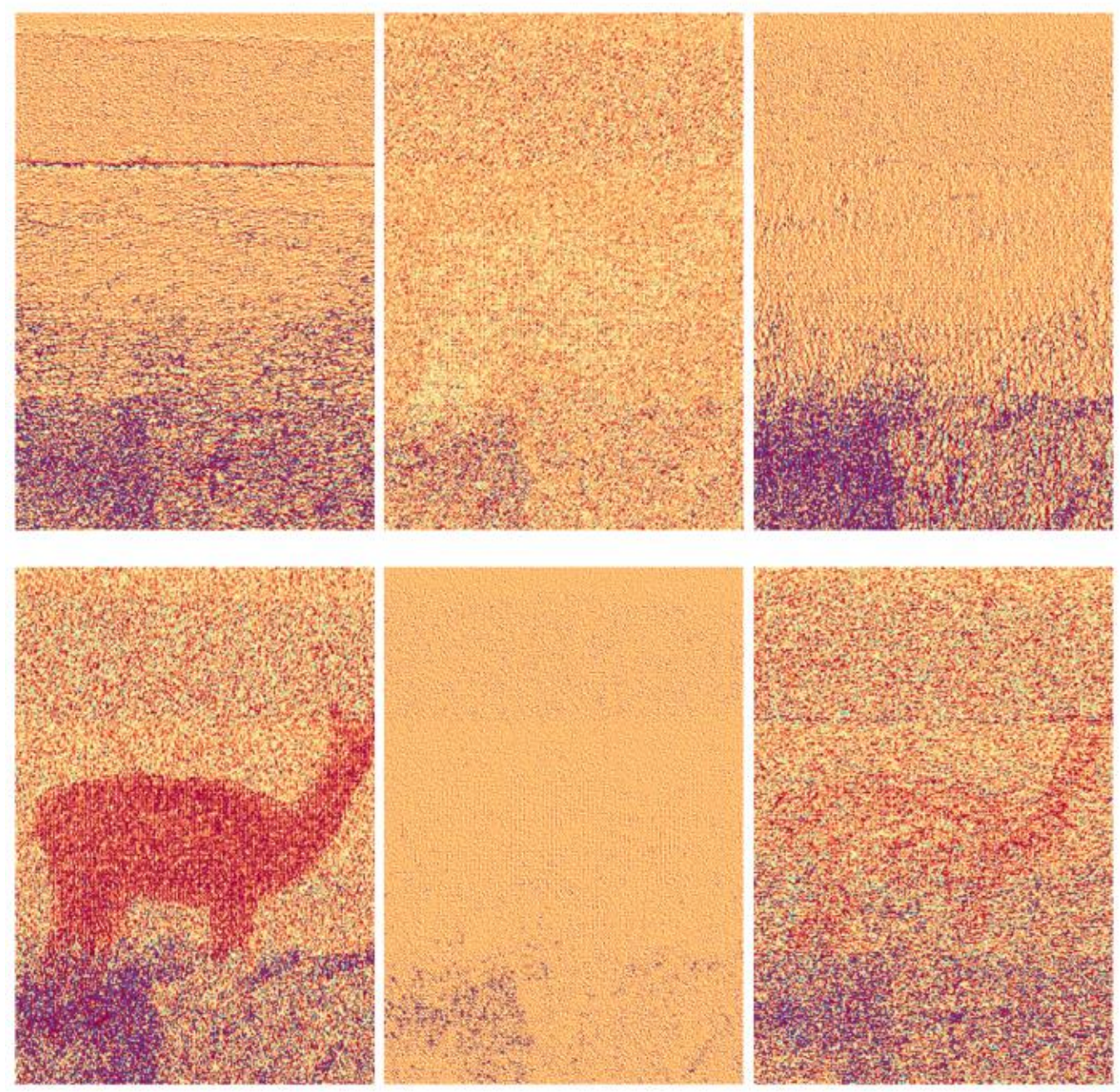

Fig. 9. Imaginary coefficients of DC-WT of the original image. On the first row are the first 3 bands and on the second row the rest of the last 3 bands

So, for e.g. if we are going to analyze band 2, real components (assuming some enhancements are done to properly display the image), the input image for the rest of algorithm should be something like:

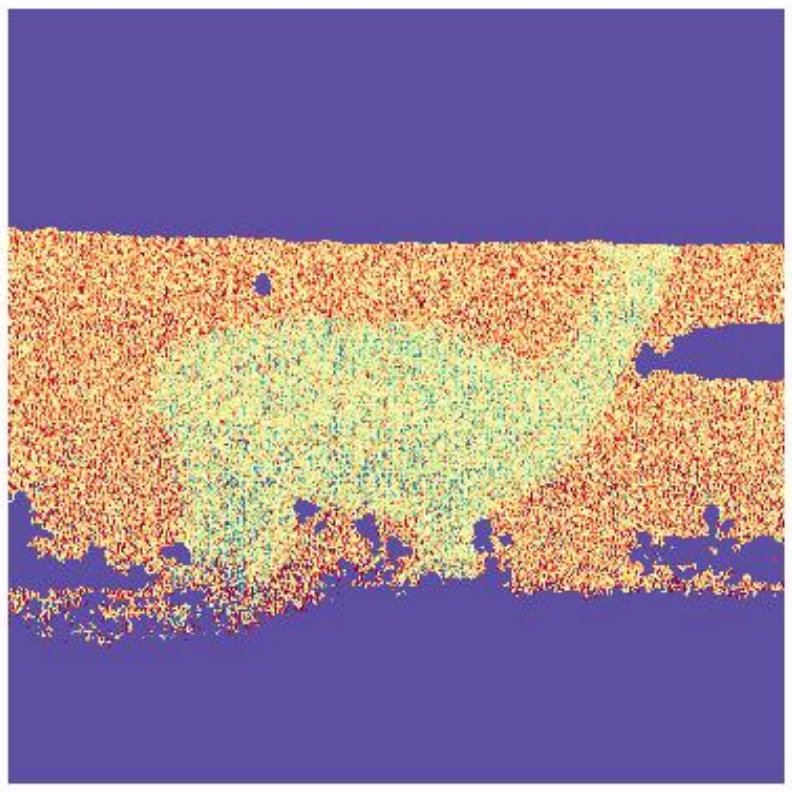

Fig. 10. Band 2 Real values for the largest segment. Further on, only the values inside the mask will be used for processing.

As a side note, to improve perceptibility in the above images, the images are enhanced by normalizing and clamping values outside the interval (-1, 2), and applying the Matplotlib Spectral color scheme.

### B. *Wavelet Band filtering*

As stated above for one segment, the method will extract 12 different matrices. Some of the matrices might not contain relevant features for the given image. For e.g. in Fig. 8 one can noticed that only bands 1,3,5 contain some visible traces, while the rest of bands (2,4,6) do not contain visible traces. To determine which bands, contain relevant information for each band, it is split into patches and patches mean is computed. If there is a higher variance among the mean patches inside the segmented area the band will be further processed, otherwise it will be skipped.

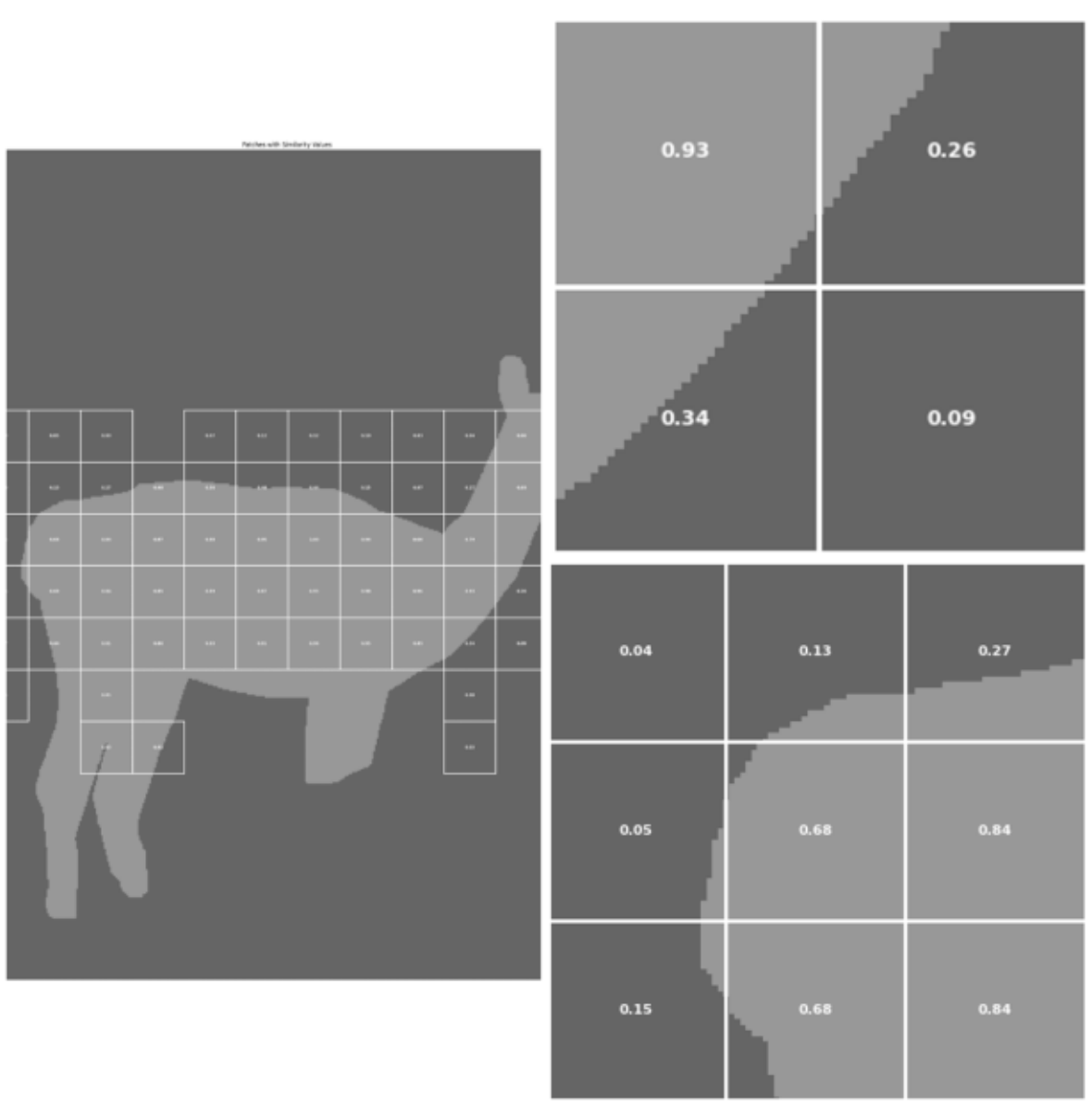

Fig. 11. Noise variance (patch means) inside band 2 real coefficients

### C. *Band enhancements*

After the bands with inconsistencies have been identified, the next step consists in enhancing the altered region to be able to detect it easier. For this several filtering are employed:

- Median Filtering: Removes impulsive noise from individual bands while preserving important structural details.
- SMFR: Isolates and enhances fine details and edges within the wavelet bands, reducing noise and highlighting features.
- Wiener Filtering: Adaptively reduces noise in each band, enhancing signal fidelity and preserving details.
- Median Modified Wiener Filtering: Combines the noise reduction and edge-preserving benefits of median and Wiener filtering, providing robust enhancement across the wavelet bands.

Median filtering is a non-linear process commonly used to reduce noise, particularly salt-and-pepper noise, while preserving edges in the image. The formula for median filtering is: Median Filtered Image (i, j) = median ({I (i+k, j+l)}), fork, $l \in [-N/2, N/2]$, where N is the kernel size. The advantages of median filtering include its effectiveness in removing salt-and-pepper noise and its ability to preserve edges better than linear filters. The Small Median Filter Residue (SMFR) technique isolates fine details and noise by computing the difference between the original image and its median-filtered version, followed by another median filtering. This process can be described with the following steps: First, the filtered image is obtained using a median filter. Next, the first difference is calculated as the original image minus the filtered image. Then, a second median filtering is applied to this first difference to obtain the second filtered difference. Finally, SMFR is computed as the first difference minus the second filtered difference. The advantages of SMFR include its ability to enhance fine details and edges while reducing noise and retaining important features. Wiener filtering is a technique that minimizes the mean square error between the estimated and true image by adapting to the local variance of the image. The Wiener filter can be mathematically represented as: Wiener Filtered Image (i, j) = $\mu + (\sigma^2 - \nu^2)/\sigma^2 \times (I(i, j) - \mu)$, where $\mu$ and $\sigma^2$ are the local mean and variance, and $\nu^2$ is the noise variance. The advantages of Wiener filtering include its adaptive noise reduction capabilities and its ability to preserve signal details in regions of low noise. Median modified Wiener was proposed in [17]. This filtering method combines the strengths of median filtering and Wiener filtering to effectively reduce noise and enhance the image. The process begins with the application of a median filter to remove impulsive noise, followed by Wiener filtering to adaptively reduce the remaining noise. The combined formula can be expressed as: First, apply median filter: Median Filtered Image (i, j) = median ({I (i+k, j+l)}). Then, apply Wiener filter to the median-filtered image: Wiener Filtered Median Image (i, j) = $\mu + (\sigma^2 - \nu^2)/\sigma^2 \times$ (Median Filtered Image (i, j) - $\mu$). The advantages of this technique include its effectiveness in removing various types of noise and its ability to preserve edges and fine details

## D. Clustering

In the proposed method, because we consistently operate on a single segment color during band decomposition, we assume that there will always be a maximum of three different clusters (2 while there is only one-color segment for entire area – e.g. image of a sky). One cluster represents the ignored area (black in our case), while another represents high-level features that are evenly distributed within the segmented color. If the segmentation results in two different types of texture, it indicates that the segment contains forged areas.

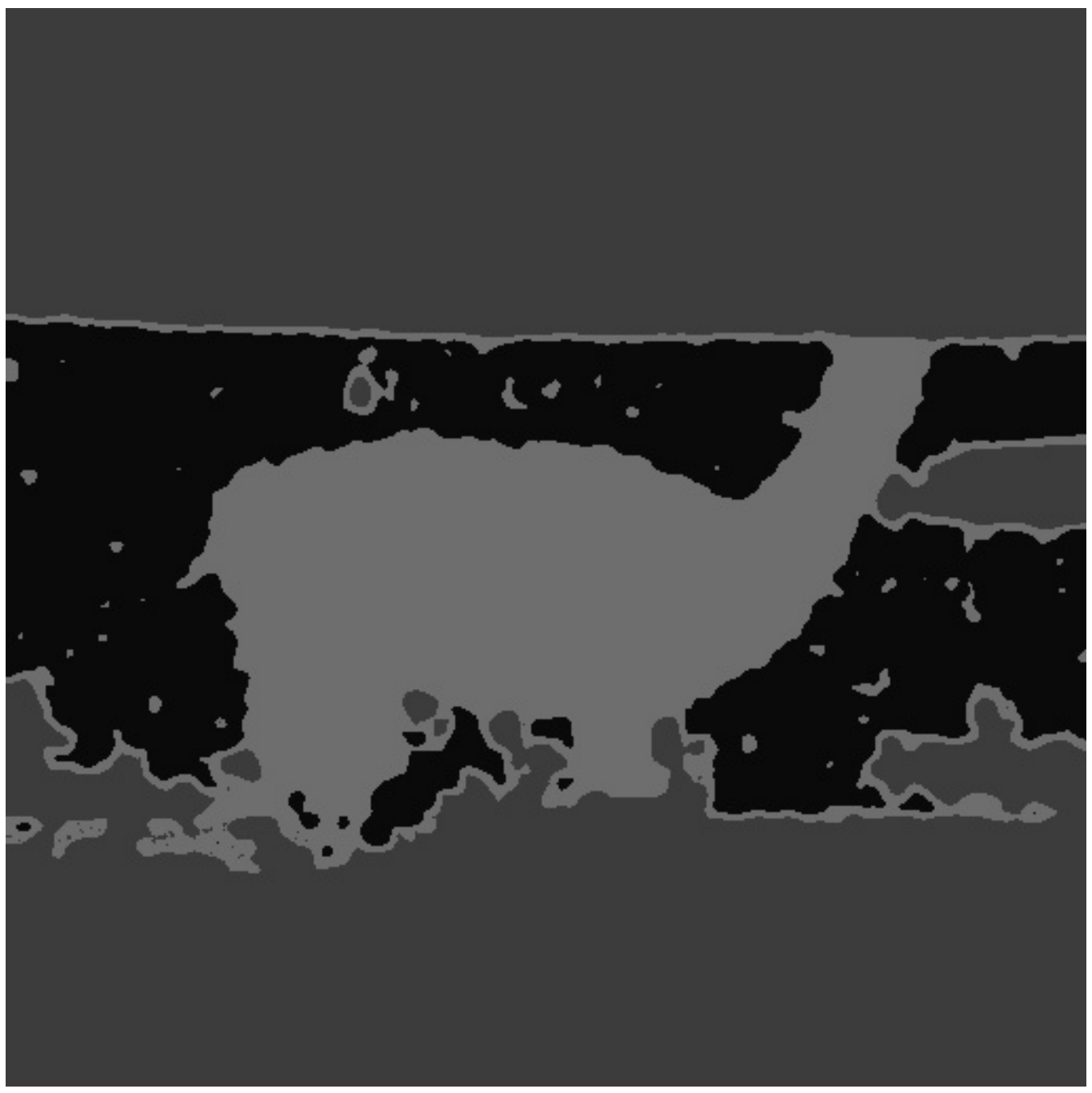

Fig. 12. Result after clustering. The dark gray area represents the outside of mask. The darker area and the lighter area represent the 2 distinct regions insisde the mask

## E. Forged area selection

All the steps thus far essentially divide the color segmentation mask into two different regions. The final step is to determine which of these sub-regions is forged and which is not. To make this distinction, the noise variance method is utilized [15]. Noise variance is calculated on the original band matrix. The band matrix is divided into three distinct regions, and the noise variance is computed for each region as well as for the entire band matrix. Using these four values, the outlier, which indicates the forged area, is identified.

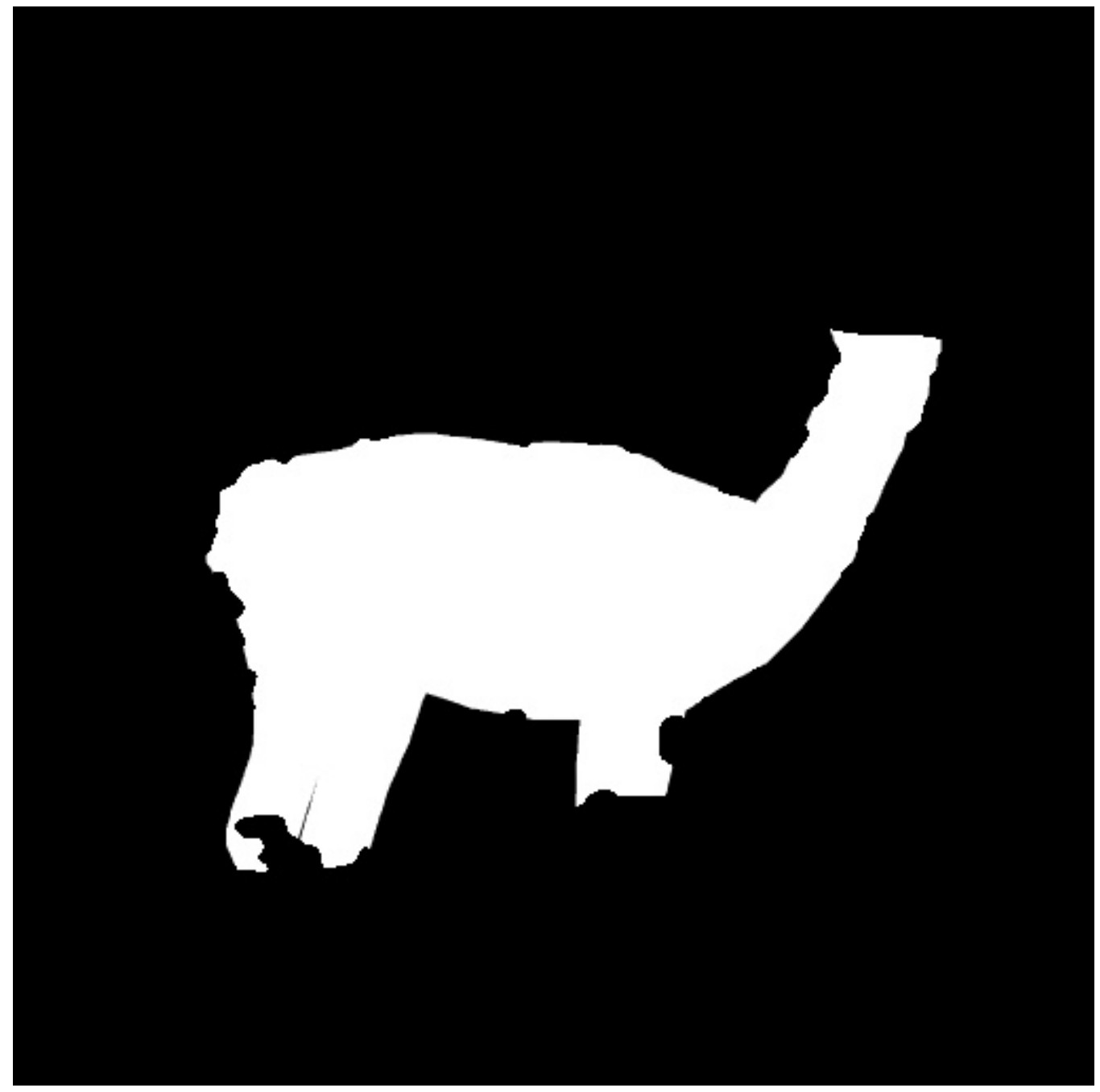

Fig. 13. Final mask per band

# V. EXPERIMENTAL RESULTS

## A. Method parameters

This chapter presents the experimental evaluation of the proposed inpainting forgery detection method employing DT-CWT and HFS. The settings used are the following: for DT-CWT the Kingsbury Q-shift filters are used; for normalization the log normalization is employed (actually symmetric log – similar to what Matplotlib is doing),; for HFS segmentations a SLIC value of 32 is used; for patch mean computation a patch size of 32 is used no padding with a tolerance of 0.5; for K-Means the cluster is either set to 3 or to 2 (when there is only on segment in the entire image – for e.g. a clear blue sky or a green field); for noise variance between areas a patch size of 8 is used with a padding of 3.

## B. Dataset

This section examines the results of various inpainting detection techniques applied to different image inpainting processes, with a primary focus on the dataset. While other forgery datasets like Casia and MFC exist, they do not specifically address image inpainting or object removal. Previous analysis highlighted that each forgery detection method often requires its own specific dataset for training and testing. To address this, Google's Open Images Dataset V7 (released in October 2022) was used [18]. From this dataset, 1k images (increased from 400 from the original paper) with segmented masks were manually selected, ensuring each image used only one mask and avoided highly texturized areas, as inpainting methods struggle with such regions. The selected images were limited to a maximum size of 1024x1024 pixels. The dataset included a variety of object sizes to be removed, from small to large. Additionally, to improve inpainting results, a dilation with a 5x5 kernel was added to masks that were close to the image borders. The complete dataset, including the original images, masks, inpainting results, and forgery detection results, can be accessed at https://github.com/jmaba/ImageInpaitingDetectionAReview

### C. Evaluation metrics

This section evaluates the performance of the proposed method for detecting tampered regions. The algorithm's effectiveness is assessed by comparing its output binary mask (indicating tampered pixels) with ground truth masks of tampered regions in the test images. Two key metrics, commonly used in image segmentation tasks, are employed to quantify performance: accuracy, and intersection over union (IoU). These metrics are calculated at the pixel level for each test image, providing a thorough evaluation of the algorithm's ability to accurately identify tampered pixels:

- **Accuracy:** Measures the proportion of correctly classified pixels, including both tampered and untampered pixels. It reflects the algorithm's overall effectiveness in distinguishing between the two classes. A high accuracy value indicates strong performance in identifying both true positives and avoiding false positives/negatives.
- **Intersection over Union (IoU)**: Combines precision and recall into a single metric. It represents the overlap area between the predicted tampered mask and the ground truth mask, divided by the total area of their union. A high IoU value signifies a good balance between precision and recall, indicating the algorithm accurately identifies most tampered pixels while minimizing false positives.

### D. Results

The proposed method was evaluated using an extended version of the publicly available inpainting forgery detection dataset. The results highlight the method's effectiveness in identifying tampered regions across a broader range of image content and inpainting techniques. We have selected several images, including the originals, masks, and inpainting results, for demonstration. For these we have applied the detection methods proposed in [19] hereafter referred to as IID, [20] hereafter referred to as PSCCNET respectively [21] named FOCAL.

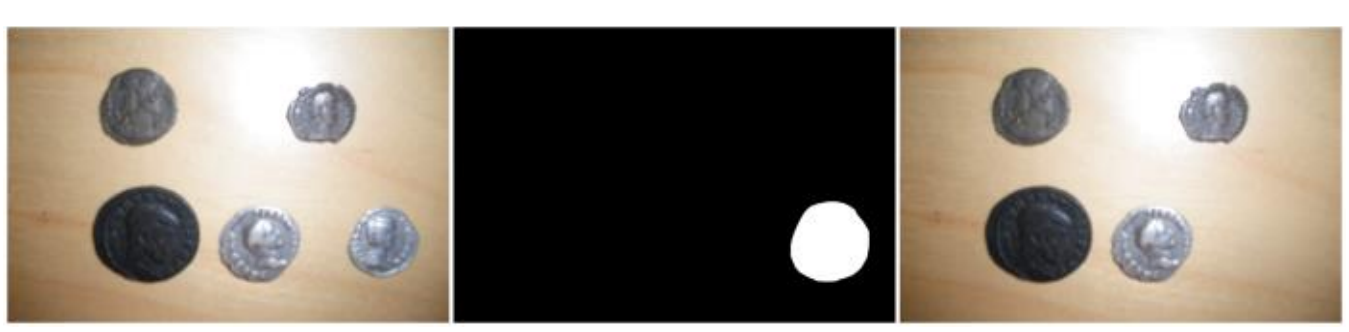

Fig. 14. Original image on left, in the middle the mask of the object to be removed (also used for computing metrics), the altered image

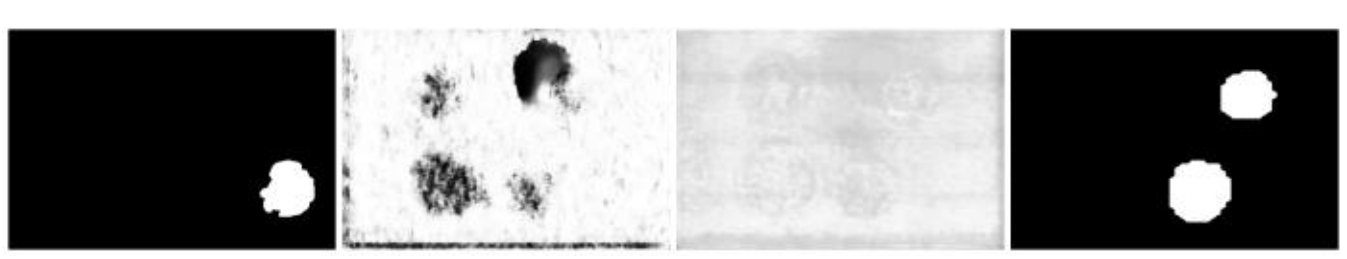

Fig. 15. First image is our generated mask, second is the mask output by IID, third is the mask output by PSCNET and last one is mask generated by FOCAL

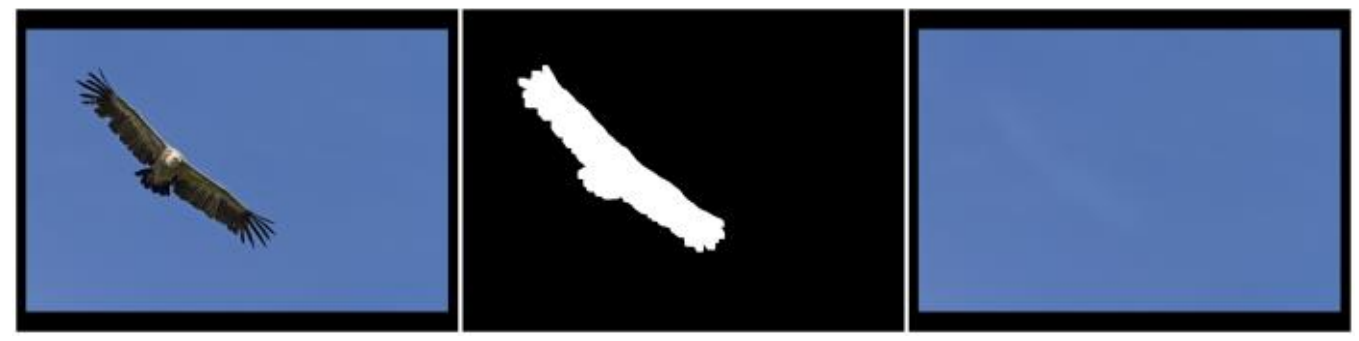

Fig. 16. Original image on left, in the middle the mask of the object to be removed (also used for computing metrics), the altered image

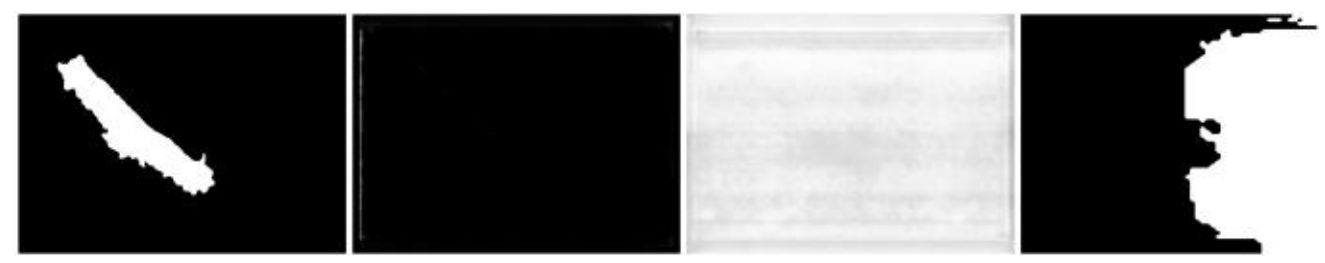

Fig. 17. First image is our generated mask, second is the mask output by IID, third is the mask output by PSCNET and last one is mask generated by FOCAL

TABLE II. TABLE TYPE STYLES

| Detection | LAMA inpainting method | | |
|---|---|---|---|
| | *Accuracy* | *Recall* | *iOu* |
| Proposed method | 0.93 | 0.89 | 0.67 |
| FOCAL | 0.75 | 0.15 | 0.10 |
| IID | 0.49 | 0.73 | 0.26 |
| FOCAL | 0.58 | 0.60 | 0.21 |

From the above images and from table results, both our method and IID yield good results. FOCAL on the other hand yields false positive results, while PSCCNET, depending how the mask interval is analyzed, might be able to detect some portions of the mask. Analysis of the IID detection method reveals high recall values, indicating successful identification of most forged images. However, accuracy and Intersection over Union (IoU) metrics are low, suggesting that the method frequently misclassifies entire images as forged rather than precisely locating the tampered regions. These conclusions can be seen on the entire dataset, not particularly to the LAMA inpainting method. While IID detection shows decent performance in some cases, it generally exhibits lower IoU scores compared to ours. IID detection tends to have high recall scores, indicating its ability to capture a large portion of the actual forged regions. The high recall values usually obtained by IID algorithms are because the method yields not a binary mask for inpainting detection, but rather for each pixel a confidence score. In some situations, the method seems to falsely attribute altered values to mostly all the pixels inside the image. IID accuracy and IoU scores vary across different inpainting methods. The PSCCNET detection method demonstrates relatively moderate performance across all inpainting methods, with accuracy, recall, and IoU scores falling between those of FOCAL and our method. The proposed method demonstrated promising results for inpainting forgery detection across various inpainting techniques. It achieved consistently high accuracy, suggesting its effectiveness in distinguishing tampered from original regions. However, the effectiveness of all methods, including ours, varied depending on the specific inpainting technique used.

## VI. CONCLUSIONS

Future studies should incorporate a broader range of inpainting methods to provide a more comprehensive assessment of how well the proposed and baseline methods perform across various scenarios. Expanding the dataset to include newer inpainting techniques would enhance the robustness of the inpainting forgery detection evaluation framework. While accuracy is an important metric for evaluating detection performance, the consistently high accuracy values achieved by the proposed method might indicate a bias towards correctly classifying untampered pixels. Future research could employ additional metrics, such as recall and IoU, which focus more on accurately identifying tampered regions.

An IoU greater than 0.5 is generally considered a good detection result, but the observed variations in recall and IoU across different inpainting techniques highlight potential limitations. The methods, including the proposed one, might not be equally effective in detecting artifacts introduced by various inpainting techniques. Additionally, processing time, particularly for the noise estimation method and clustering, is a significant limitation that needs to be addressed.

Another limitation is that some configuration values were obtained through extensive testing. These parameters may vary for other types of inpainting, suggesting the need for a machine learning-based approach to automatically adjust parameters for different scenarios. Employing adaptive methods or leveraging machine learning for parameter tuning could improve the robustness and effectiveness of forgery detection across various inpainting techniques. Furthermore, the proposed method is highly susceptible to different types of post-processing, such as resizing, blurring, and noise addition. These post-processing techniques can significantly affect the detection performance, highlighting the need for more resilient detection methods capable of handling various image manipulations.

## REFERENCES

[1] A. Criminisi, P. Pérez, and K. Toyama, “Region filling and object removal by exemplar-based image inpainting,” *IEEE Transactions on Image Processing*, vol. 13, no. 9, pp. 1200–1212, Sep. 2004, doi: 10.1109/TIP.2004.833105.

[2] K. Papafitsoros, C. B. Schoenlieb, and B. Sengul, “Combined First and Second Order Total Variation Inpainting using Split Bregman,” *Image Processing On Line*, vol. 3, pp. 112–136, Jul. 2013, doi: 10.5201/IPOL.2013.40.

[3] J. Yu, Z. Lin, J. Yang, X. Shen, X. Lu, and T. S. Huang, “Generative Image Inpainting with Contextual Attention,” *Proceedings of the IEEE Computer Society Conference on Computer Vision and Pattern Recognition*, pp. 5505–5514, Jan. 2018, doi: 10.1109/CVPR.2018.00577.

[4] C. Corneanu, R. Gadde, and A. M. Martinez, “LatentPaint: Image Inpainting in Latent Space with Diffusion Models,” *Proceedings - 2024 IEEE Winter Conference on Applications of Computer Vision, WACV 2024*, pp. 4322–4331, 2024, doi: 10.1109/WACV57701.2024.00428.

[5] R. Rombach, A. Blattmann, D. Lorenz, P. Esser, and B. Ommer, “High-Resolution Image Synthesis with Latent Diffusion Models,” *Proceedings of the IEEE Computer Society Conference on Computer Vision and Pattern Recognition*, vol. 2022-June, pp. 10674–10685, 2022, doi: 10.1109/CVPR52688.2022.01042.

[6] I. W. Selesnick, R. G. Baraniuk, and N. G. Kingsbury, “The dual-tree complex wavelet transform,” *IEEE Signal Process Mag*, vol. 22, no. 6, pp. 123–151, 2005, doi: 10.1109/MSP.2005.1550194.

[7] T. Julliand, V. Nozick, and H. Talbot, “Image Noise and Digital Image Forensics,” *Lecture Notes in Computer Science (including subseries Lecture Notes in Artificial Intelligence and Lecture Notes in Bioinformatics)*, vol. 9569, pp. 3–17, 2016, doi: 10.1007/978-3-319-31960-5_1.

[8] H. Zeng, A. Peng, and X. Lin, “Exposing image splicing with inconsistent sensor noise levels,” *Multimed Tools Appl*, vol. 79, no. 35–36, pp. 26139–26154, Sep. 2020, doi: 10.1007/S11042-020-09280-Z/METRICS.

[9] Q. Liu, H. Li, and Z. Liu, “Image forgery localization based on fully convolutional network with noise feature,” *Multimed Tools Appl*, vol. 81, no. 13, pp. 17919–17935, May 2022, doi: 10.1007/S11042-022-12758-7/METRICS.

[10] Z. Zhang *et al.*, “Noise and Edge Based Dual Branch Image Manipulation Detection,” *ACM International Conference Proceeding Series*, pp. 963–968, May 2023, doi: 10.1145/3603781.3604221.

[11] J. Chen, X. Liao, W. Wang, Z. Qian, Z. Qin, and Y. Wang, “SNIS: A Signal Noise Separation-Based Network for Post-Processed Image Forgery Detection,” *IEEE Transactions on Circuits and Systems for Video Technology*, vol. 33, no. 2, pp. 935–951, Feb. 2023, doi: 10.1109/TCSVT.2022.3204753.

[12] M. Gardella, P. Muse, J. M. Morel, and M. Colom, “Noisesniffer: A Fully Automatic Image Forgery Detector Based on Noise Analysis,” *Proceedings - 9th International Workshop on Biometrics and Forensics, IWBF 2021*, May 2021, doi: 10.1109/IWBF50991.2021.9465095.

[13] R. Suvorov *et al.*, “Resolution-robust Large Mask Inpainting with Fourier Convolutions,” *Proceedings - 2022 IEEE/CVF Winter Conference on Applications of Computer Vision, WACV 2022*, pp. 3172–3182, 2022, doi: 10.1109/WACV51458.2022.00323.

[14] A. Kuznetsova *et al.*, “The Open Images Dataset V4: Unified image classification, object detection, and visual relationship detection at scale,” *Int J Comput Vis*, vol. 128, no. 7, pp. 1956–1981, Nov. 2018, doi: 10.1007/s11263-020-01316-z.

[15] “[PDF] An Efficient Statistical Method for Image Noise Level Estimation | Semantic Scholar.” Accessed: Jul. 11, 2024. [Online]. Available: https://www.semanticscholar.org/paper/An-Efficient-Statistical-Method-for-Image-Noise-Chen-Zhu/3924f6b1ab44a35370a8ac8e2e1df5d9cd526414

[16] M. M. Cheng *et al.*, “HFS: Hierarchical feature selection for efficient image segmentation,” *Lecture Notes in Computer Science (including subseries Lecture Notes in Artificial Intelligence and Lecture Notes in Bioinformatics)*, vol. 9907 LNCS, pp. 867–882, 2016, doi: 10.1007/978-3-319-46487-9_53/FIGURES/7.

[17] C. V. Cannistraci, F. M. Montevecchi, and M. Alessio, “Median-modified Wiener filter provides efficient denoising, preserving spot edge and morphology in 2-DE image processing,” *Proteomics*, vol. 9, no. 21, pp. 4908–4919, Nov. 2009, doi: 10.1002/PMIC.200800538.

[18] A. Kuznetsova *et al.*, “The Open Images Dataset V4: Unified image classification, object detection, and visual relationship detection at scale,” *Int J Comput Vis*, vol. 128, no. 7, pp. 1956–1981, Nov. 2018, doi: 10.1007/s11263-020-01316-z.

[19] H. Wu and J. Zhou, “IID-Net: Image Inpainting Detection Network via Neural Architecture Search and Attention,” *IEEE Transactions on Circuits and Systems for Video Technology*, vol. 32, no. 3, pp. 1172–1185, Mar. 2022, doi: 10.1109/TCSVT.2021.3075039.

[20] X. Liu, Y. Liu, J. Chen, and X. Liu, “PSCC-Net: Progressive Spatio-Channel Correlation Network for Image Manipulation Detection and Localization,” *IEEE Transactions on Circuits and Systems for Video Technology*, vol. 32, no. 11, pp. 7505–7517, Nov. 2022, doi: 10.1109/TCSVT.2022.3189545.

[21] H. Wu, Y. Chen, and J. Zhou, “Rethinking Image Forgery Detection via Contrastive Learning and Unsupervised Clustering,” Aug. 2023, Accessed: Sep. 28, 2023. [Online]. Available: https://arxiv.org/abs/2308.09307v1